\documentclass{article}
\usepackage{arxiv}

\usepackage{graphicx}
\usepackage{amsmath}
\usepackage{amssymb}
\usepackage{amsfonts}       
\usepackage[x11names]{xcolor}
\usepackage{booktabs} 
\usepackage{natbib}
\usepackage{appendix}
\usepackage{multirow}
\usepackage[margin=2cm]{caption}

\usepackage[pagebackref=true,breaklinks=true,colorlinks,bookmarks=false]{hyperref}

\begin{document}

\definecolor{atomictangerine}{rgb}{0.8, 0.2, 0.1}
\definecolor{turq}{rgb}{0.0, 0.5, 0.5}
\definecolor{darkturq}{rgb}{0.0, 0.4, 0.4}
\definecolor{bright}{rgb}{0.8, 0.1, 0}
\definecolor{darkgray}{gray}{0.3}
\definecolor{mahogany}{rgb}{0.6, 0.05, 0.05}
\definecolor{pink}{rgb}{1,0.05,0.6}
\definecolor{olive}{rgb}{0.537, 0.627, 0.318}
\definecolor{green}{rgb}{0.22, 0.463, 0.114}
\definecolor{grey}{rgb}{0.4, 0.4, 0.4}
\definecolor{blue}{rgb}{0.435, 0.659, 0.863}
\definecolor{pink}{rgb}{0.761, 0.482, 0.627}
\definecolor{red}{rgb}{0.6, 0, 0}
\definecolor{myblue}{rgb}{0.3,0.05,0.9}
\newcommand\gal[1]{\textcolor{red}{[GC: #1]}}
\newcommand\yuval[1]{\textcolor{teal}{\textbf{YA:} #1 }}
\newcommand\galo[1]{\textcolor{cyan}{\textbf{GO:} #1 }}
\newcommand\gilad[1]{\textcolor{blue}{\textbf{GV:} #1 }}

\newcommand{\tildeapprox}{{\raise.17ex\hbox{$\scriptstyle\sim$}}}
\newcommand{\ignore}[1]{}
\newcommand{\loss}{\mathcal{L}}
\renewcommand{\eqref}[1]{Eq.~\ref{#1}}
\newcommand{\appref}[1]{Appendix \ref{#1}}
\newcommand{\figref}[1]{Fig.~\ref{#1}}
\newcommand{\figureref}[1]{Figure~\ref{#1}}
\newcommand{\tabref}[1]{Tab.~\ref{#1}}
\newcommand{\tableref}[1]{Table.~\ref{#1}}
\newcommand{\secref}[1]{Sec.~\ref{#1}}
\newcommand{\reals}{\mathbb{R}}
\newcommand{\wvec}{\mathbf{w}}
\newcommand{\PSGSM}{PSST Gumbel softmax}
\newcommand{\PSMS}{PSST Multinomial}
\newcommand{\PSST}{PSST}
\newcommand{\STGSM}{Straight-through Gumbel Softmax}
\newcommand{\STmulti}{straight-through multinomial}
\newcommand{\reinforce}{{\sc{reinforce}}}
\newcommand{\etal}{\textit{et al}.}
\newcommand{\Luo}{Luo \etal 2018}
\newcommand{\cider}{CIDEr}

\title{Cooperative image captioning}
 
\author{Gilad Vered\\
Bar-Ilan University\\
\and
Gal Oren \\
Bar-Ilan University\\
\and
Yuval Atzmon \\
Bar-Ilan University, \\ NVIDIA research\\
\and
Gal Chechik \\
Bar-Ilan University, \\ NVIDIA research\\
{\texttt{gal.chechik@biu.ac.il}}
}

\maketitle
\begin{abstract}
    When describing images with natural language, the descriptions can be made more informative if tuned using downstream tasks. This is often achieved by training two networks: a "speaker network" that generates sentences given an image, and a "listener network" that uses them to perform a task. Unfortunately, training multiple networks jointly to communicate to achieve a joint task, faces two major challenges. First, the descriptions generated by a speaker network are discrete and stochastic, making optimization very hard and inefficient. Second, joint training usually causes the vocabulary used during communication to drift and diverge from natural language. 
    
    We describe an approach that addresses both challenges. We first develop a new effective optimization based on partial-sampling from a multinomial distribution combined with straight-through gradient updates, which we name \textbf{\PSST} for \textit{Partial-Sampling Straight-Through}. Second, we show that the generated descriptions can be kept close to natural by constraining them to be similar to human descriptions. Together, this approach creates descriptions that are both more discriminative and more natural than previous approaches. Evaluations on the standard COCO benchmark show that \PSMS{} dramatically improve the recall@10 from $60\%$ to $86\%$ maintaining comparable language naturalness, and human evaluations show that it also increases naturalness while keeping the discriminative power of generated captions.
\end{abstract}

\section{Introduction}
\label{sec:introduction}
Describing images with natural language is a key step for developing automated systems that communicate with people. The complementary part of this human-machine communication involves networks that can understand natural descriptions of images. Both of these tasks have been studied intensively, but mostly as two separate problems, \textit{image captioning} and \textit{image retrieval}. It is therefore natural to "close the loop" and seek to jointly train networks that can cooperatively communicate about visual content in natural language. 

Training multiple networks to communicate has been studied recently in the context of visual dialogues \cite{das2017learning,Das_2017_CVPR}. There, a series of sentences is passed back-and-forth between learning agents. Here we take a step back and focus on a \textit{single} transmission between a "speaker network" and a "listener network", We seek to develop the building blocks of trainable communication by training both speaker and listener to communicate effectively with natural language.

What should be the properties of such natural communication? Good visual descriptions should obey two competing objectives. First, a  description should be natural and fluent, using well-formed and meaningful sentences, so they can be communicated to people. Second, a description should be image-specific and informative, capturing the relevant elements of an image that make it unique.

Aiming to address these two objectives, \cite{luo2018discriminability} trained a speaker network together with a pre-trained listener network that evaluates the discriminative power of the descriptions by detecting its corresponding image among similar distractor images. Similarly, \cite{dai2017diverse} described a GAN approach that evaluates both naturalness and discriminability.
For our purpose, we also wish to obtain a listener network that can perform well with the speaker. 

Unfortunately, training a speaker network together with a listener network leads to major challenges: language-drift and optimization.  First, when the listener and speaker can tune their communication, the resulting language typically \textbf{drifts away}, loses its original semantic meaning, and makes it confusing for communicating with people. For instance, networks may assign a new meaning (blue) to a common word (red), or code highly-specific information within a single symbol (``field'' becomes synonym for a ``giraffes near a tree'') \citep{kottur2017natural}.

Second, training end-to-end speaker-listener systems requires to optimize through an intermediate communicating layer which is discrete and typically stochastic. Standard backpropagation of gradients cannot be applied to such layers 
\citep{bengio2013estimating}, and alternative methods are often complex or slow to converge \citep{tucker2017rebar,williams1992simple}. Because of these limitations, previous approaches like in  \citep{vedantam2017context,luo2018discriminability} avoided end-to-end training or obtained limited quality captions \citep{dai2017diverse}. 

The current paper addresses these two challenges. First, we show that keeping the discriminative captions close to human-generated captions, is sufficient for maintaining fluent and well-formed language while providing enough flexibility such that captions are discriminative. Second, we develop a new effective optimization procedure for jointly training a cooperative speaker-listener combination. It based on partial-sampling from a multinomial distribution combined with straight-through gradient updates, which we name \textbf{\PSST{}} for \textit{Partial-Sampling Straight-through}. It can be applied with straight-through optimization, or with \STGSM{}. \PSST{} is very simple to implement, and robustly outperforms all baselines we compared with. 

This paper makes the following novel contributions (1) A new and simple partial-sampling procedure for optimizing through discrete stochastic layers, directly applicable to generating discriminative language. (2) New state-of-the-art results on MS COCO discriminative captioning, improving recall from  $\tildeapprox 60\%$ to $\tildeapprox 86\%$ for similar naturalness evaluated using CIDEr and human evaluation. 
(3) Systematic evaluation of all the leading approaches for optimization through stochastic layers, using a unified captioning benchmark. (4) An evaluation scheme that explicitly quantifies the full curve of naturalness-vs-discriminability, instead of a one-dimensional metric. 


\setlength{\baselineskip}{13pt}

\section{Discriminative captioning}
\label{sec:discaption}
\begin{figure}[ht]
    \centering
    \includegraphics[width=0.50\linewidth]{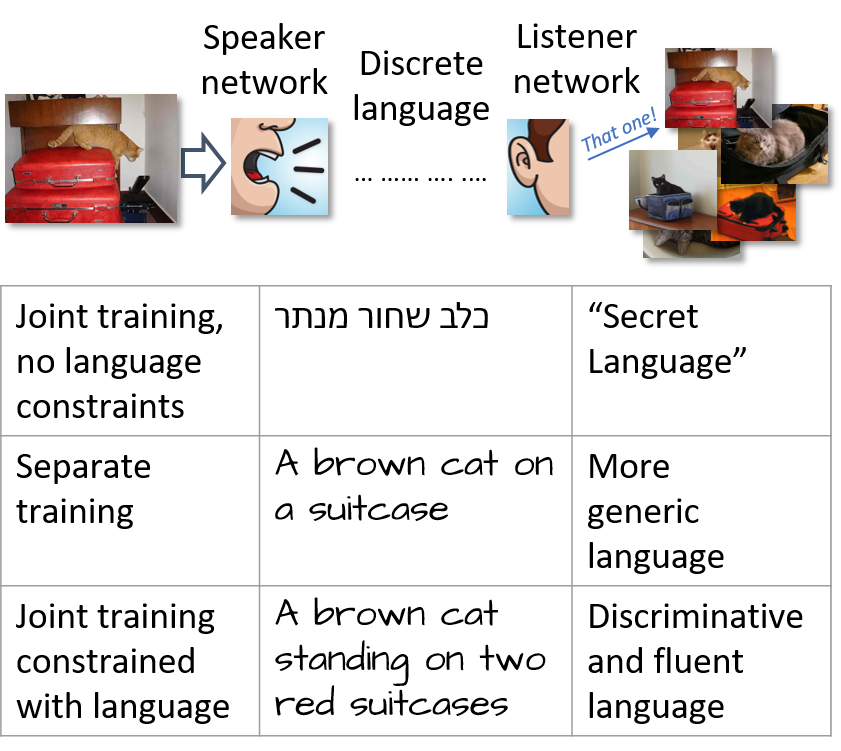}
    \caption{\textit{The challenge of training two agents to communicate about an image. \textbf{Top row:} When a speaker network is trained jointly with a listener network, the communication drifts away from natural language, if unconstrained. The resulting language no longer maps to standard English terms. \textbf{Middle row}: If the agents are trained separately, descriptions become less specific, because agents cannot count on the other side to "understand" them. \textbf{Bottom row}: Training both networks jointly while restricting communication to be close to natural language can yield descriptions that are more discriminative while maintaining intelligibility.}}
    \label{fig:teaser}
\end{figure}

In our setup of discriminative captioning, two networks cooperate to communicate the content of a given image (\figref{fig:teaser}). The first,  \textbf{speaker}, network is given an image and produces a series of discrete tokens that describe the image \textit{in natural language}. Each token is represented by a 1-hot vector from a predefined vocabulary. The second, \textbf{listener}, network takes this series of tokens and uses it to find the input image among a set of distractor images. The two networks share a common goal: communicate such that the listener identifies the image that the speaker described. 

In this setup, the speaker network is trained to focus on the unique features of an image that would allow the listener to detect it among distractors. However, the specific distractor images are not available to the speaker as an explicit context, a setup that was studied in   \citep{vedantam2017context,jhamtani2018learning}. Importantly, both networks share a common objective, and their interaction defines a cooperative game. 
This is fundamentally different from adversarial approaches GANs \citep{dai2017diverse}.

We address this task by training the speaker network jointly with the listener network. When considering the objective function of this joint optimization, it must contain two complementing components. First, as a \textbf{discriminability loss} the objective contains the loss suffered by the listener when detecting the target image among distractors. Since natural language is far from optimal for this task, the networks can find other communication schemes that drift away from natural language. To keep the communication interpretable to people, we add a second component to the objective, a \textbf{naturalness loss}, aimed to keep sentences natural (similar to \cite{luo2018discriminability}). To this end, we add to the loss a measure of similarity between the generated caption and human-created captions for that image. Specifically, we experimented below with CIDEr as a similarity measure. The overall objective is a weighted combination of the naturalness loss and the discriminability loss weighted by a trade-off parameter $\lambda$.

It is important to realize that when training networks, the two components of the objective compete with each other. One reason is that automatic measures of naturalness quantify how well a caption matches a pre-defined set of human-generated captions, that were not created for a discriminative task.

\section{Related work}
\label{sec:related}
Image captioning has been studied intensively since encoder-decoder models were introduced \citep{xu2015show}. Large efforts have been invested in making captions more natural and diverse. For example, \cite{dai2017diverse} used conditional GANs to train a caption generator to improve fidelity, naturalness,  and diversity. Using GANs allows avoiding the hard challenge of defining explicit language naturalness loss. Instead, the discriminator can receive fake or incorrect captions or images as negatives. \cite{chen2018} used a conditional GAN with two discriminators, a CNN and an RNN. \cite{dai2018neural} further used a hierarchical compositional model over captions to increase diversity and naturalness. More related to the optimization techniques discussed in this paper, \cite{shetty2017speaking} trained an adversarial network using a straight-through Gumbel approach. As we discuss below, training cooperative agents allows using more effective optimization techniques compared to training GANs, because the generator is allowed to provide any useful information to the (cooperative) discriminator. Specifically, during training, the speaker can represent generated captions differently than human captions.  

Beyond the naturalness of communication, several studies looked into the problem of generating captions that allow discriminating an image from other similar images.  \cite{vedantam2017context} showed how captions can take into account a distractor image at inference time and create a caption that discriminates a target image from a distractor image. A similar approach was taken earlier by \cite{andreas2016reasoning}. \cite{hexiang2018bison} recently described a dataset that contains pairs of closely similar images, that can be used as hard-negatives for evaluating image retrieval tasks. 

Most relevant to the current paper is the work of \cite{luo2018discriminability}. They showed how to use a pre-trained listener network for increasing caption discriminate power. However, to avoid language drift, they kept the listener network fixed, rather than training jointly with the speaker  network. 

Several authors studied the properties of languages that are learned when agents communicate in visual tasks, \cite{kottur2017natural,bouchacourt2018agents,lee2018emergent,lazaridou2016multi}. The current paper purposefully focuses on keeping the language close to natural, rather than study properties or emergent language.

\begin{figure}
    \centering
    \includegraphics[width=0.98\linewidth]{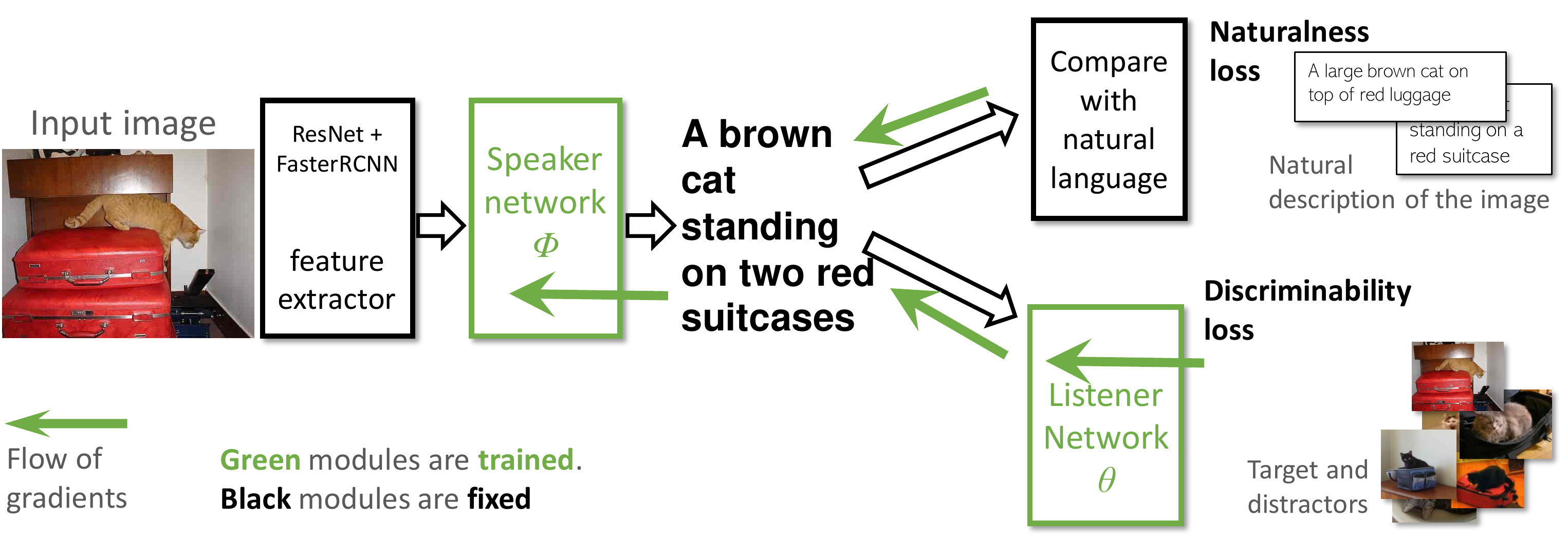}
    \caption{\textit{The architecture of our system. The speaker network and the listener networks are trained jointly, by passing gradients through the text layer. The loss contains two components, that are linearly weighted with a hyper parameter tuned on the validation set. \textbf{Naturalness loss:} A measure of agreement between a generated caption and a set of predefined, ground-truth captions for that image. Those captions need not be discriminative. The experiments below used CIDEr. \textbf{Discriminative loss:} Measures how well a listener can identify the image among a set of 127 randomly-chosen distractor images.}}
    \label{fig:architecture}
\end{figure}

\section{Optimizing discrete stochastic layers}
\newcommand\ft{{f_{\theta}}}
\newcommand\ftw{{f_{\theta}(\wvec)}}
Joint training of two networks communicating through a language layer is equivalent to training a network that has an intermediate layer that is discrete and stochastic.  
We describe below the main existing methods for this problem, but first define formally the learning setup.

In our model (\figref{fig:architecture}), caption generation is treated as a stochastic process. At each step, $t=0,\ldots,T$ the caption generator (the speaker) outputs a distribution over a vocabulary of words $p_{\phi}(w_t|I, w_0,...,w_{t-1})$. This distribution depends on the input image $I$ and the previous terms in the sentence and is parametrized by the deterministic parameters $\phi$. We therefore treat the output of the speaker network $s_\phi(I)$ as a random sequence $W$ with a distribution $p_\phi(\wvec)$ over all word sequences $\wvec$. From that distribution, one specific sequence is sampled and passed to the listener. Given this sampled word sequence $\wvec=w_0,\ldots,w_T$, the listener network, parametrized by $\theta$, makes a prediction $\hat{y} = \ftw$ = $\ft(s_{\phi}(I))$, and suffers a loss $l(y, \hat{y}; \theta)$. Our goal is to propagate the gradient of that loss, first to update the parameters of the listener $\theta$ and then through the stochastic layer to update the parameters of the speaker $\phi$.

Propagating the losses to the parameters of the listener poses no special problems to the computation graph since the function $f_{\theta}$ implemented by the listener network is deterministic and differentiable (almost everywhere). This is also true for propagating the gradients back through the sequence of terms in a sentence, which can be done using standard ``back-propagation through time''.

\newcommand\ltheta{{L_{\theta}}}
\newcommand\lthetaw{{l_{\theta}(\wvec)}}
\newcommand\lthetawi{{l_{\theta}(w_i)}}
\newcommand\nabphi{{\nabla_{\phi}}}

However, tuning the parameters of the speaker network poses two problems: Terms are discrete - hence non-differentiable, and their selection is stochastic - again non-differentiable. Computation in stochastic neural networks can be formalized using stochastic-computation graphs (SCGs) \citep{Schulman2015GradientEU}. In our case, we view the computation graph as including a single stochastic computing node, corresponding to the random sequence $W$. We think about the listener network as providing the speaker with a loss $\lthetaw$ for every sentence $\wvec$, and our goal is to minimize the expected loss
\begin{equation}
    \min_{\phi} L(\theta, \phi)) = \min_{\phi} E_{\phi(\wvec)} \left[\lthetaw\right].
\end{equation}  
The gradient of this objective w.r.t. $\phi$, the parameters of the speaker, is
\begin{equation}
    \label{eq:grad}
    \nabphi \int p_\phi(\wvec)\lthetaw d\wvec = \int \nabphi p_\phi(\wvec)\lthetaw d\wvec .
\end{equation}
Since it does not have a form of an expectation, it cannot be directly estimated  efficiently by sampling.

Several solutions were proposed for this problem, including a score-based function approach \citep{williams1992simple}, and a Gumbel soft-max approach \citep{maddison2017concrete, JangGuPoole17}. For completeness, we first describe these approaches shortly and then discuss in detail a new and simple partial-sampling approach from a multinomial distribution, combined with straight-through gradient updates.

\subsection{Score-function estimators}
\label{sec:reinforce}
The \reinforce{} algorithm \citep{williams1992simple}, also known as score-function estimator \citep{Fu2006,Glynn1990} is usually described in the context of reinforcement learning. There, an agent aims to maximize its reward by choosing the best action for a given state according to a policy. In our context of discriminative image captioning, the state is determined by the input image and the preceding words, the set of possible actions are the set of words that can be emitted as the chosen word, and the reward is (minus) the loss imposed by the listener.

\newcommand\pphix{{p_\phi(x)}}
Using the identity $\nabphi \pphix = \pphix \nabphi \log \pphix$, the gradient w.r.t. $\phi$ in \eqref{eq:grad} can be rewritten as (\cite{maddison2017concrete}):
$\nabphi L(\theta, \phi) = E_{p_{\phi}(\wvec)}\left[ \lthetaw \nabphi \log p_{\phi}(\wvec)\right]$. This formulation allows us to estimate the expectation by sampling and computing the empirical mean over samples. 
$   \nabphi L(\theta, \phi) \approx \frac{1}{n} \sum_{i=1}^{n} \lthetawi \nabphi \log p_{\phi}(w_i). $

Unfortunately, while this estimator of the gradient is unbiased, it is known to have large variance, which leads to slow convergence. Several techniques have been proposed for reducing the variance, while maintaining an unbiased estimator \citep{Gu2016,tucker2017rebar,grathwohl2017backpropagation, mnih2016variational}. Unfortunately, these techniques tend to be complex to implement and analyze, hence their adoption is still limited. 

\subsection{\STGSM{} }
\label{sec:Gumbel}
A second approach to optimize the stochastic discrete layer is using \textit{\STGSM{} } \citep{JangGuPoole17,maddison2017concrete}. It combines three steps which we review shortly. First, to handle stochasticity, the computation graph is reparameterized, allowing to propagate gradients through deterministic paths only. Second, the Gumbel max process is used for sampling from a pre-determined distribution and the Gumbel distribution is relaxed using a Gumbel softmax. Finally, a "straight-through" trick is used to propagate gradients back. We now explain these three components in more details.

\paragraph{Reparametrizing the stochastic computation graph.}
The reparametrization trick,  \citep{kingma2013auto,rezende2014stochastic}, is based on restructuring the computation graph such that the stochastic node is moved to a side branch of the graph that is outside the gradient-propagation path. Gradients can then be propagated through the main path that is deterministic. Reparametrization allows to sample from a known fixed distribution $q(z)$, and then use a deterministic function $g_\phi(z)$ to transfer the sampling to specific distribution $p_\phi(\wvec)$. Following \cite{maddison2017concrete} we write
$ L(\theta,\phi) = {\mathbb{E}}_{W{\sim}_{p\phi(w)}}{[\ftw]} = {\mathbb{E}}_z{_\sim}_{q_(z)}{[f_\theta(g_\phi(Z))]}$, and $ \nabphi L{(\theta,\phi)} = 
   {\mathbb{E}}_z{_\sim}_{q(z)}{[\nabla_\wvec f_\theta(\wvec)\nabla_\phi(g_\phi(Z)]} $. 
As can be seen above, $f_\theta(\wvec)$ does not have to be differentiable w.r.t. $\phi$. 

\paragraph{Gumbel Max.}
Given the above reparameterization, we can sample from a given discrete distribution as follows \citep{Gumbel1954statistical}. First, sample from a uniform distribution $u \sim U(0,1)$ and compute ${g=-\log(-\log(u))}$. To sample from a desired categorical distribution $p_i$, use
$ z={one\_hot}{\left({\arg\max_i}[g_i+ \log p_i]\right)}.$
\citep{JangGuPoole17}.
To allow propagating gradients, relax the max using a soft-max, yielding the continuous \textbf{Gumbel Softmax} distribution \citep{JangGuPoole17}. The Gumbel softmax distribution approximates a categorical distribution
$ y_i=\frac{\exp{(\log({p_i})+g_i})/\tau}{\sum_{j=1}^{k}\exp{((\log(p_j)+g_j)/\tau)}}$. The parameter $\tau$ controls the softmax temperature.: As it approaches to 0, sample become closer to one-hot vectors, but the variance of the gradients is large. when $\tau$ is large, the variance is small but samples are smoother and as a result also away from the original categorical distribution. 

\paragraph{Straight-through estimators.} 
Sampling with Gumbel softmax produces "soft" outputs, rather than one-hot. To generate specific sentences, we are forced to commit to a specific instance of that distribution, namely, to select the most likely symbol based on the stochastic distribution by taking the argmax of the distribution. Unfortunately, this argmax operation is not differentiable.
To address this, the \textbf{\STGSM{} }  \citep{JangGuPoole17} sets the forward pass to pass the argmax over the Gumbel-Softmax distribution, while in the backward pass, gradients are computed as if the full continuous distribution was passed. This estimator is related to the binary straight-through approach described in \cite{bengio2013estimating}. It is usually biased because there is a mismatch between how parameters are updated during the backward pass and the actual activation in the forward pass. 

\section{Partially-sampled Straight through}
\label{sec:PSMS}
The Gumbel softmax approach described above succeeds to bypass the issue of optimization with stochastic units. However, for applications like captioning where a discrete output is required, it resorts to using a straight-through approach.

This approach has several disadvantages. First, the forward pass is stochastic, adding inherent \textbf{variance} to the optimization process, and conveying less information per sample than passing the full continuous distribution. Indeed, presenting the same input to the network leads to different estimates of the gradients. Second, the straight-through estimator is also \textbf{biased}, because the estimates of the gradients are computed as if the full distribution was passed. It would have been preferable to pass the full distribution without sampling. Unfortunately, at test time we must produce discrete word selections to generate specific sentences. 

We propose a simple-to-implement procedure we call \textit{partial-sampling straight-through} (\PSST{}). During training, we pass the full continuous distribution for a fraction $\rho$ of the samples. 
In other words, for $\rho$ of the samples, the stochastic and discrete units are practically replaced by a deterministic and differentiable  variable. 
The remaining $1-\rho$ of the samples can be optimized either by passing a sampled one-hot from the multinomial distribution, which we call \textit{\PSMS{}} or using Gumbel Softmax, which we call \textit{\PSGSM{}}. 

In the extreme case of  $\rho=0$, the speaker always operates as a sampler. The \PSMS{} optimization can be viewed as a multinomial version of the binary straight-through estimator of \cite{bengio2013estimating}. In the other extreme case of $\rho=1$, the speaker operates as a deterministic mapper and outputs a set of dense multinomial distributions.   

This approach has several advantages. First, for $\rho$ of the samples, the estimator of the gradient is exact, because computation is deterministic, therefore reducing the overall bias and variance of gradient estimation, by a factor of $\rho$. At the same time, for $1-\rho$ of training images, the downstream listener network does experience stochastic variations, sparse sentences are represented as 1-hot vectors, and it  learns to classify them correctly. This allows it to correctly handle one-hot samples that are observed during the test phase. We find empirically that this approach is highly effective and robust with respect to the value of $\rho$.

Partial sampling takes advantage of the cooperative nature of the speaker-listener relations. Unlike GAN training  \citep[e.g.][]{dai2017diverse}, where the generator works hard not to reveal any information that may give away its generated captions, the speaker in cooperative games has an explicit aim to convey as much information as possible to the listener. Specifically, during training, it is allowed to represent generated captions as continuous distributions, which look very different than human-created captions, and would be easily discriminated by GANs. More generally, the fundamental differences in the "game matrix" of communicating agents, cooperative vs competitive, are important to consider when developing joint optimization schemes.

\subsection*{Our Approach}
\label{sec:our-approach}
We summarize our approach to discriminative captioning. 
We train jointly a speaker and listener networks, aiming to minimize a loss that has two components:  A \textit{discriminative loss} $l^{disc}$, and a \textit{naturalness loss} $l^{nat}$. 
\begin{align}
    \label{eq:loss}
    \min_{\phi,\theta}  \lambda l^{disc}(\wvec, I) + (1-\lambda) l^{nat}(\wvec)
\end{align}
where $\phi$ are the parameters of the speaker network and $\theta$ are the parameters of the listener network. For the naturalness loss, we use  CIDEr \citep{vedantam2015cider}, $l^{nat}(\wvec) = - CIDEr(\wvec)$. For the discriminative loss $l^{disc}$ we use the sum of two hinge losses: one for selecting the correct image among a batch of distractor images,  and a second for selecting the correct caption among a batch  distractor captions as in \citep{faghri2017vse++}:
\begin{align}
    l^{disc}(\wvec,I) &= \max\left[0, 1 - \Phi_{\theta}(\wvec, I) + \Phi_{\theta}(\wvec',I) \right] \\ \nonumber
    &+ \max\left[0, 1 - \Phi_{\theta}(\wvec, I) + \Phi_{\theta}(\wvec, I') \right] ,
    \end{align}
where $\wvec'$ is the hardest negative caption among candidate captions, $I'$ is the hardest negative image, and $\Phi$ is the cosine similarity over the embedding of the image and captions.
Instead of fixing a single value of $\lambda$, we compute the full curve that captures the trade-off between discriminative and natural descriptions, obtained by optimizing the model with varying values of $\lambda$.

To optimize $l^{disc}(\wvec)$, we applied the \PSMS{} procedure as described \secref{sec:PSMS}. To optimize $l^{nat}(\wvec)$ we cannot use \PSMS{} because $l^{nat}(\wvec)$ requires sparse descriptors as input. Instead, we elected to optimize $l^{nat}(\wvec)$ with \reinforce{} because preliminary experiments showed that its performance was comparable to the other approaches discussed above (see \figref{fig:recall_vs_cider}). 
 


\section{Experiments}
\label{sec:experiments}
We evaluate our approach two image captioning benchmark datasets:  COCO  \citep{lin2014microsoft}, and Flickr30k \citep{young2014image}. 

The COCO dataset has \tildeapprox{123K} images, where each image is annotated with 5 human-generated captions.  For a fair comparison with previous work, we used the same data split as in \citep{vedantam2017context,luo2018discriminability}, assigning $\tildeapprox{113K}$, $5K$ and $5K$ images for training, validation and test splits.
We used the vocabulary of 9487 words as in \cite{luo2018discriminability}.

The Flickr30K dataset has \tildeapprox{31K} images, annotated with 5 human-generated captions for a total of \tildeapprox{159,000} captions. We used the split as in \cite{karpathy2015deep} assigning $29K$, $\tildeapprox1K$ and ${1K}$ images for train, validation and test splits. The vocabulary contains words that appeared more than 5 times in the annotated captions, yielding a vocabulary of $7K$ words. Captions that were longer than 16 words were clipped to this length.

\subsection{Experimental setup and hyper parameters}
On COCO, to allow easy comparison with the most relevant baselines, we followed the setup proposed by \cite{luo2018discriminability} whenever possible (code provided by the authors of \citep{luo2018code}). 
On Flickr30K, to pre-train the listener with tested the following hyper parameters batch size in (\{64, 128\}), learning rate in (\{1e-3, 5e-4, 1e-4\}), and decay rate every 15 or 30 epochs. We found that the parameters yielded best recall scores were learning-rate of $5e-4$, learning-rate decay every $15$ epochs and batch size of $64$. We pre-trained the listener for 30 epochs.

For pre-training the speaker with MLE, we tested the following hyper-parameters. batch sizes in (\{64, 128\}), learning rate in (\{1e-3, 5e-4, 1e-4\}), and decay rate every 15 or 30 epochs. The parameters that yielded best CIDEr score were learning-rate of $1e-4$, learning-rate decay every $30$ epochs and batch size of $64$. We pre-trained the speaker for 100 epochs.

For the ST Multinomial method, We tested learning rate in (\{1e-3, 5e-4, 1e-4\}) and decay rate in (\{0.75, 0.8, 0.85, 0.9, 0.95, 0.99\}). A learning-rate of 1e-4 and  learning-rate decay rate of $0.9$ worked best. 

For PSST Multinomial, we tested learning rates in (\{5e-4 and 1e-4\}) and learning-rate decay rate of (\{0.7, 0.8, 0.9\}). As with ST Multinomial, the best parameters were learning-rate of 1e-4 and a decay rate of $0.9$. For all methods we trained models for 200 epochs.

\subsection{Automatic evaluation metrics}
\noindent\textbf{Naturalness}. At test time, the naturalness of generated captions was quantified by the common linguistic metrics: CIDEr \citep{vedantam2015cider}, BLEU4 \citep{papineni2002bleu}, METEOR \citep{banerjee2005meteor}, ROUGH \citep{lin2004rouge} and SPICE \citep{anderson2016spice}. 

\noindent\textbf{Discriminability} of generated captions was quantified by the performance of the listener network. Specifically, at test time, given an input image, the listener receives four inputs: the caption generated by the speaker, the input image, 4999 distractor captions, and 4999 distractor images. The listener ranks all images based on their compatibility with the caption (measured using the cosine similarity between the image representation and the caption). Based on this ranking, we compute the recall@k, the average detection rate at the top K. Namely, an image is considered detected if the score of the input image is ranked within the top-K scores. We report below recall@1, @5 and @10.. 

\noindent\textbf{Balancing discriminability with naturalness}. During training, We control the trade-off of caption discriminability vs naturalness by testing multiple values of the trade-off parameter ${\lambda}$ of \eqref{eq:loss}. We tested ${\lambda}$ values \{0.01, 0.005, 0.0025, 0.0016, 0.001, 0.0005\}. These values are small because the two components of the loss in \eqref{eq:loss} have different scales. We  report results for all values of ${\lambda}$.

\subsection{Human-based evaluation metrics}
\label{sec:amt}
Since the above automated metrics are limited and often fail to capture naturalness \citep{anderson2016spice}, we used human judgment to evaluate the quality of the generated captions. We evaluated both discriminability and naturalness, so we have introduced two types of tasks: seeking the best image for a caption and seeking the best caption for an image. In both cases, we used an evaluation dataset published by \cite{luo2018discriminability}, and a protocol similar to \cite{vedantam2017context}. The details of the protocol and rater instructions are given in the appendix, and shortly described in Tables \ref{tab:disc} and \ref{tab:nat}.

\subsection{Compared methods}
This paper focuses on better joint optimization of the speaker and listener. We therefore adhered to the same network architecture as previous approaches \citep{luo2018discriminability}, to keep comparisons meaningful. We compared the following six approaches.
\begin{enumerate}
    \item{{{\sc{\PSMS{}}}: Partial-Sampling Multinomial Straight Through}. The method described in section \ref{sec:our-approach}.}
    
    \item
    {{\sc{ST Multinomial} Straight-through Multinomial}}. As in \PSMS{}, but always sample from the distribution during the forward pass. This is identical to using $\rho=0$. This approach  was mentioned in passing in Section 2.2 of  \cite{JangGuPoole17}.

    \item{{\sc{\Luo}}}. The method of \cite{luo2018discriminability}. The speaker network was trained while using a "frozen" pre-trained listener network. The parameters of the speaker were trained using \reinforce{}. 
    
    \item{{\sc{\reinforce}}}. Based on \cite{williams1992simple}. The speaker and listener networks were trained alternately, where at each step one network is frozen and the other is being updated. We early-stop based on the validation-set recall.
    To reduce the variance of the estimator, we used as a baseline the score that the listener assigns to ground-truth captions for each given image. 
    
    \item {\sc{\STGSM{} }}. The method of \cite{JangGuPoole17}. During back-propagation, gradients flow through the noisy distribution of Gumbel softmax, and during the forward pass, tokens are sampled from that distribution.
    
    \item{\sc{\PSGSM{}}}. Similar to \PSMS{}, but applying the partial sampling approach to the Gumbel-softmax distribution. 
\end{enumerate}
    
Several earlier papers evaluated their methods on the same COCO split tested here. Some papers, like \cite{dai2017diverse,rennie2017self} reported lower CIDEr metrics, likely because they were aiming at diversity or other goals. Other papers did not measure caption discriminability, and reported higher CIDEr score. It is important to stress that in our setup, the naturalness metric and the discriminability metric  competing with each other because naturalness is measured in terms of matching a predefined set of captions that were not designed to be discriminative. 

\subsection{Implementation details}
\label{subsec:implementation-details}
For fair comparisons, we followed the experimental procedure of \cite{luo2018discriminability} whenever possible. When there were differences between the hyper parameters published with the online code \citep{luo2018code}, and those in the paper \citep{luo2018discriminability}, we adhered to the published code. 

\noindent\textbf{Network architectures and Image features} The listener network follows the architecture in \cite{faghri2017vse++}, and the speaker network is based on \cite{rennie2017self}. More details are provided in appendix A. Two types of features were extracted from images. First, 2048-dim vectors from the last layer of a ResNet-101 \citep{he2016deep} were used to train the listener. Second, Spatial features from the output of a Faster R-CNN \citep{ren2015faster} were used to train the speaker.

\noindent\textbf{Human-generated ground-truth captions}: We processed the human captions and used the vocabulary of \cite{luo2018discriminability}. Specifically, the maximum captions length is 16 and vocabulary size is 9487. 

\noindent\textbf{Pretraining and Joint training}:
We pre-trained the listener and the speaker as in \citep{luo2018code}, but used longer training to ensure convergence (listener for 30 epochs, speaker for 200 epochs). 

For methods with joint-training, we trained the listener and speaker jointly for another 150 epochs, at which point the recall@10 on the validation has saturated. We then selected the best model as "early stopping" based on the recall@10 on the validation set. All hyper parameters were selected using the validation set. See more implementation details in the appendix. 

\begin{figure}[h]
  \begin{center}
    \includegraphics[width=0.3\linewidth]{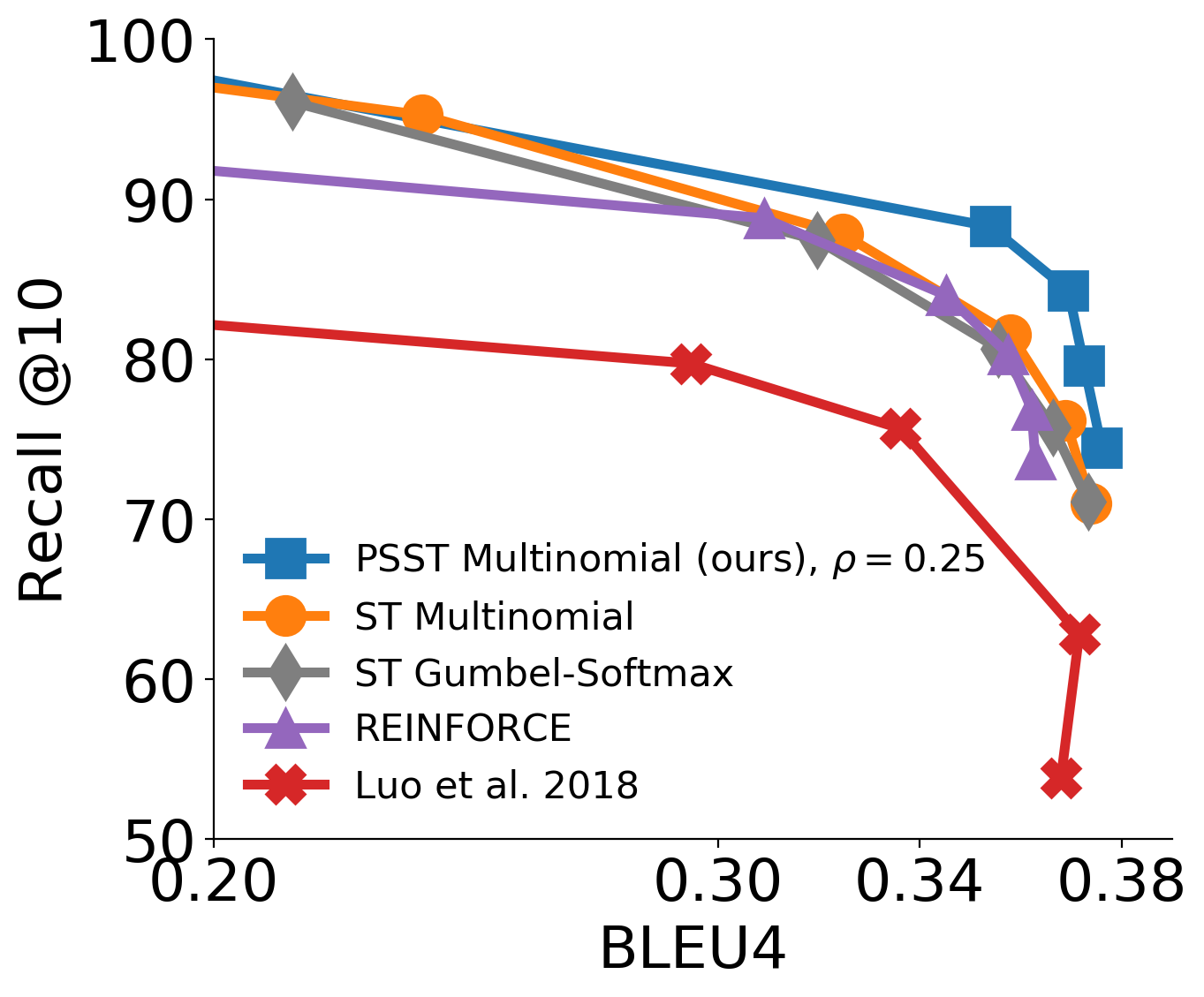}
    \includegraphics[width=0.3\linewidth]{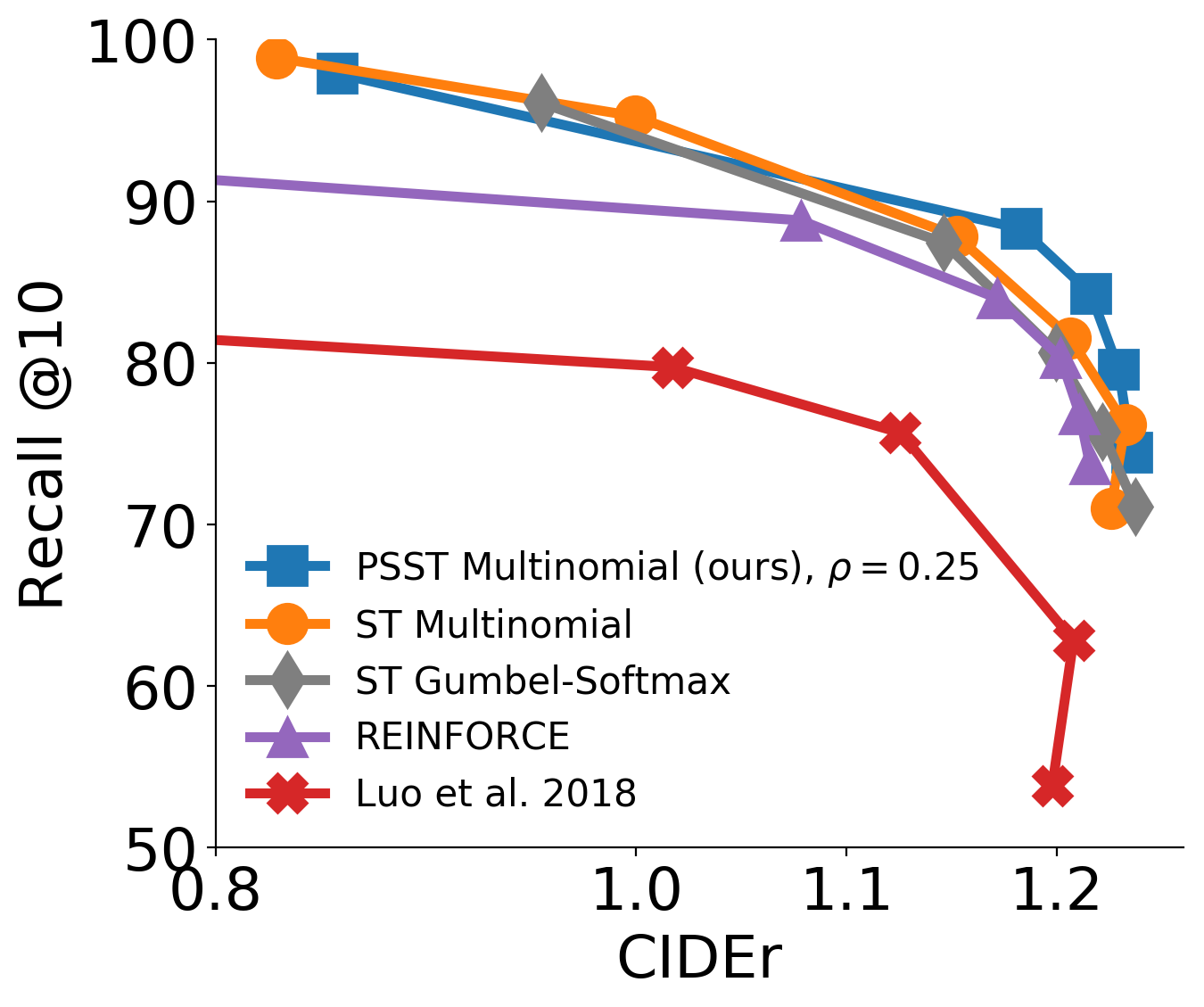}
    \includegraphics[width=0.3\linewidth]{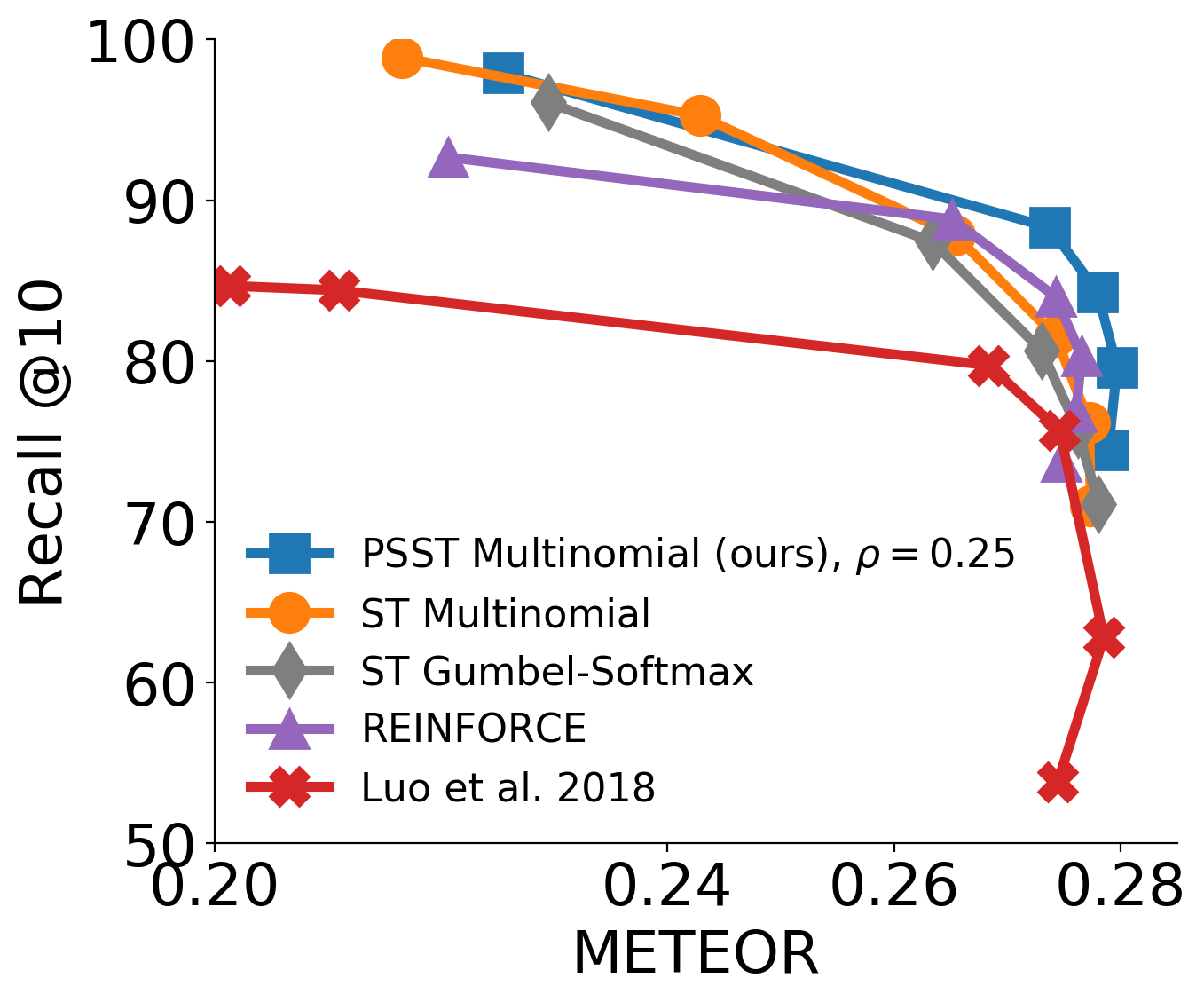}
    \includegraphics[width=0.3\linewidth]{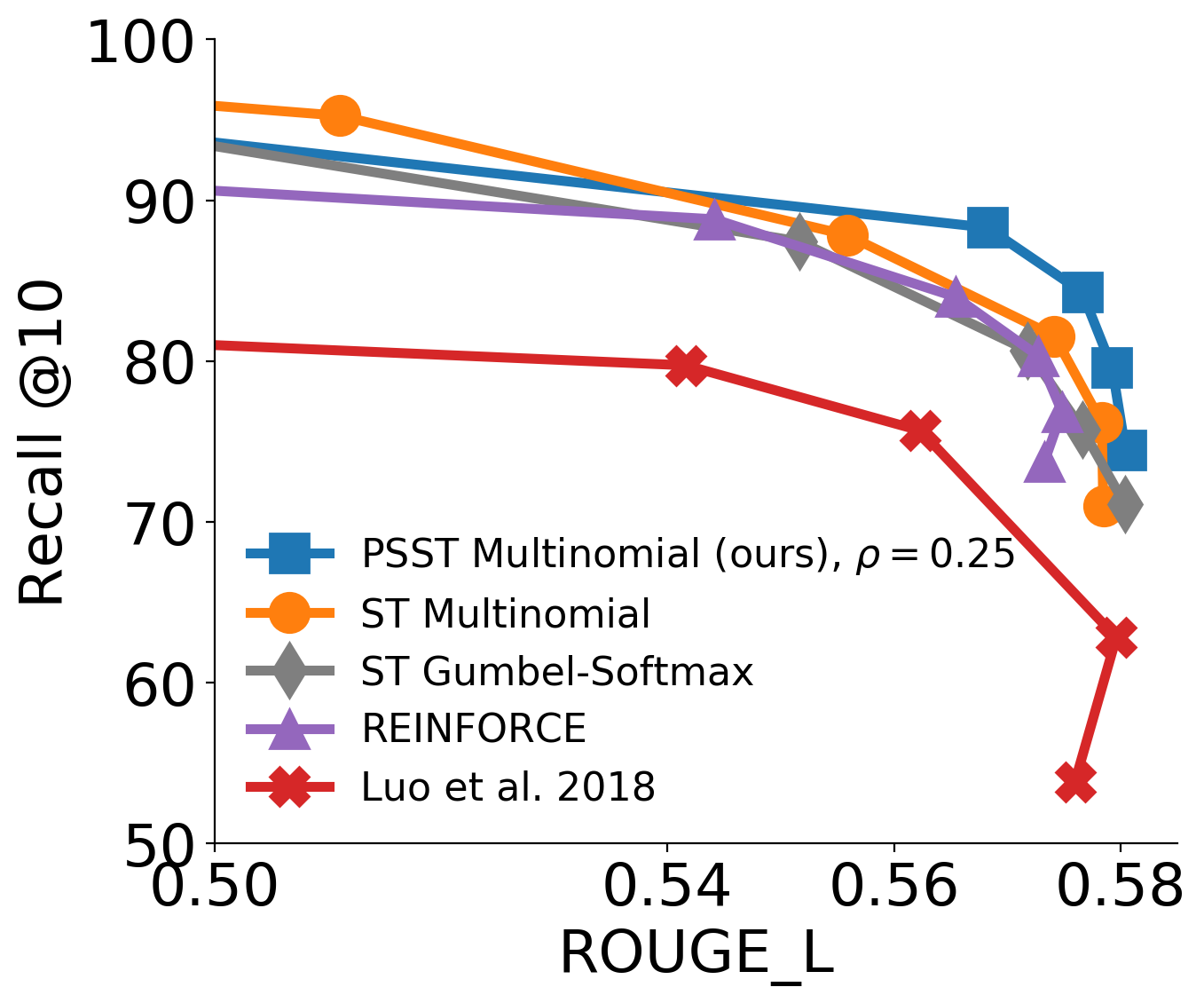}
    \includegraphics[width=0.3\linewidth]{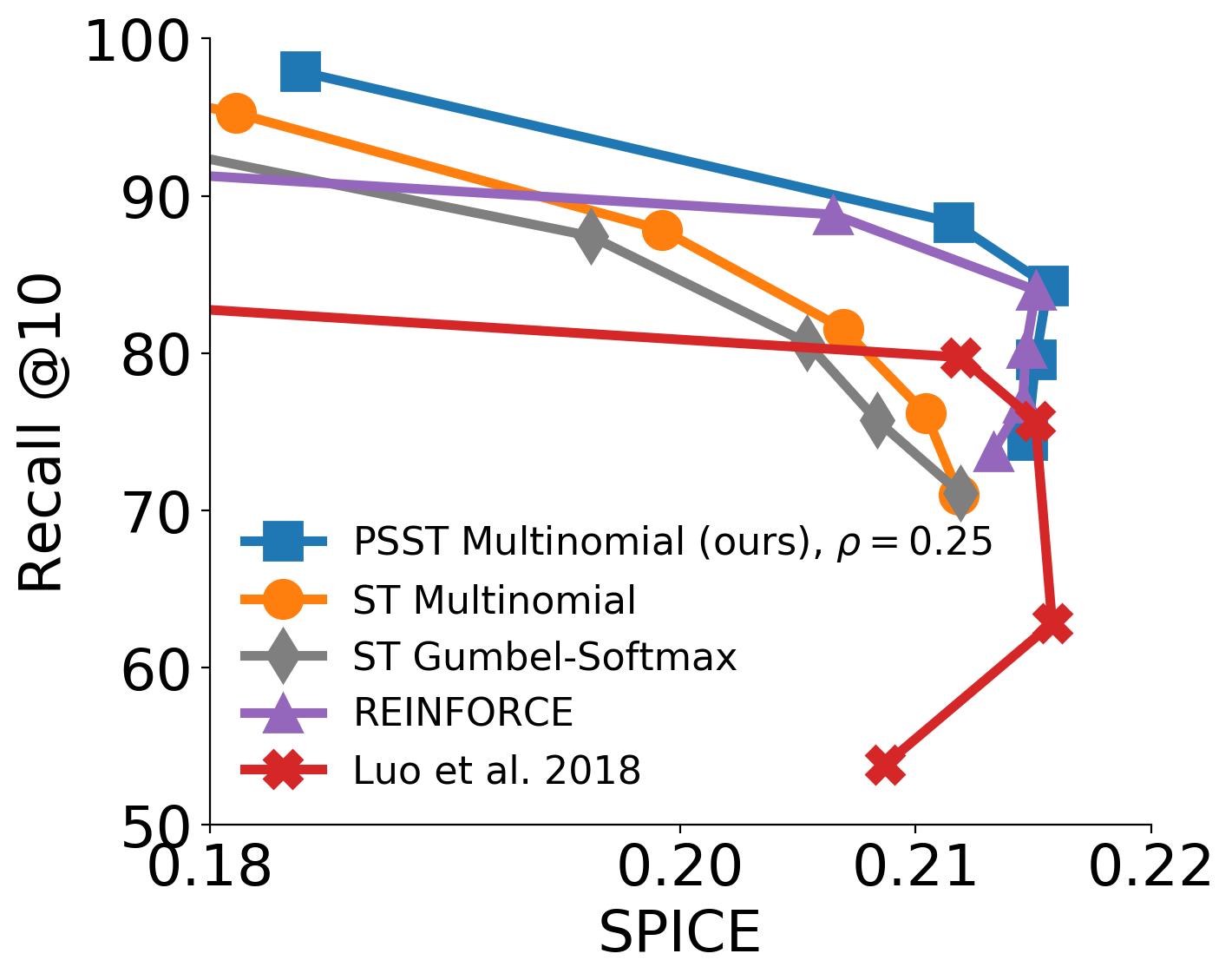}
  \end{center}
  \caption{\textit{ \textbf{Discriminability-Naturalness curves} of all compared methods on COCO. Each of the five panels shows another naturalness quantified with another language metrics: BLEU4, CIDEr, METEOR, ROUGE and SPICE. All panels have discriminability quantified using  \textit{Recall @10} of the listener. Within each panel, each curve corresponds to another method, showing a series of models trained separately with different values of the trade-off parameter $\lambda$.}}
\label{fig:recall_vs_cider}
\end{figure}

\section{Results}
We first evaluate the naturalness and discriminability of \PSST{} and the competing methods on the COCO dataset. 

\figref{fig:recall_vs_cider} depicts recall@10 as a function of five naturalness scores: {\sc{cider}\citep{vedantam2015cider}, \sc{blue4} \citep{papineni2002bleu}, \sc{meteor}, \sc{rouge}, \sc{spice}} \citep{anderson2016spice}. For each method, we trained a series of models, each with a different value of the trade-off parameter $\lambda$ (the weight of $l^{disc}$ in \eqref{eq:loss}). High values of ${\lambda}$ lead to a model that generates more discriminative captions at the expense of low language metrics, while models trained with low ${\lambda}$ generate captions with high naturalness scores but lower discriminability. The values of language metrics are provided in \tableref{table:coco_metrics}, for a fixed recall value. \PSMS achieved best scores across all five metrics. Values of Recall for a fixed \cider{} value are reported in \tabref{table:recall_fixed_cider}. Here as well, \PSMS outperforms other approaches.

Two effects are notable: \textbf{joint training}, and \textbf{partial sampling}. First, all methods that applied joint training consistently improve over separate training (red curve). Broadly speaking, all three methods, \reinforce{}, ST Gumbel softmax and ST Multinomial achieve comparable scores in the relevant region of high naturalness (BLEU4$\!\!>\!\! 0.3$ or CIDEr $\!\!>\!\! 1.1$). Second, \PSMS{} (blue curve) 
provides a significant further improvement over all baselines. 
\begin{table}[hb]
    \begin{center}
        \setlength{\tabcolsep}{2pt} 
        \begin{tabular}{lrrrrr}
            Recall@5=80\% & CIDEr & BLEU4 & METOR & ROUGE & SPICE \\
            \midrule
            \reinforce{}  &  0.902 &  0.251 &  0.247 &  0.505 &  0.189 \\
            ST Gumbel Softmax  &  1.087 &  0.288 &  0.253 &  0.528 &  0.187 \\
            ST Multinomial  &  1.106 &  0.300 &  0.259 &  0.542 &  0.194 \\
            \midrule
            PSST Gumbel Softmax (ours) &  1.109 &  0.320 &  0.263 &  0.541 &  0.205 \\
            PSST Multinomial (ours) & \bf{1.119} & \bf{0.322} & \bf{0.264} &  \bf{0.544} & \bf{0.206} \\
            \bottomrule
        \end{tabular}
    \end{center}
    \caption{\textit{\textbf{Evaluations on COCO}. {\sc{cider, blue4, meteor, rouge, spice}}, with a fixed R@5.}}
    \label{table:coco_metrics}
\end{table}
\begin{table*}[h]
    \begin{center}
        \begin{tabular}{lrrr}
        {CIDEr=1.2}&   R@1 &   R@5 &  R@10     \\
        \midrule
        Luo et al. 2018             & 16.1 &  41.5 &  55.8 \\
        \reinforce{}                & 31.3 &  67.0 &  80.5 \\
        ST Gumbel Softmax           & 31.8 &  67.3 &  80.6 \\
        ST Multinomial              & 31.9 &  67.9 &  81.8 \\
        PSST Gumbel Softmax (ours)  & 37.6 &  73.0 &  85.7 \\
        PSST Multinomial (ours)     & \textbf{38.1} &  \textbf{74.2 }&  \textbf{86.3} \\
        \bottomrule
        \end{tabular}
    \end{center}
    \caption{\textit{\textbf{Recall for a fixed CIDEr}, comparing recall for fixed values CIDEr, as extracted from \figref{fig:recall_vs_cider}. The metrics are reported on a high \cider{} operating point, showing the strong effect of joint training and the superiority of our approach. For both PSST methods, $\rho=0.25$ was used.}}
    \label{table:recall_fixed_cider}
\end{table*}

\begin{figure}[ht]
    \centering
    \includegraphics[width=0.50\linewidth]{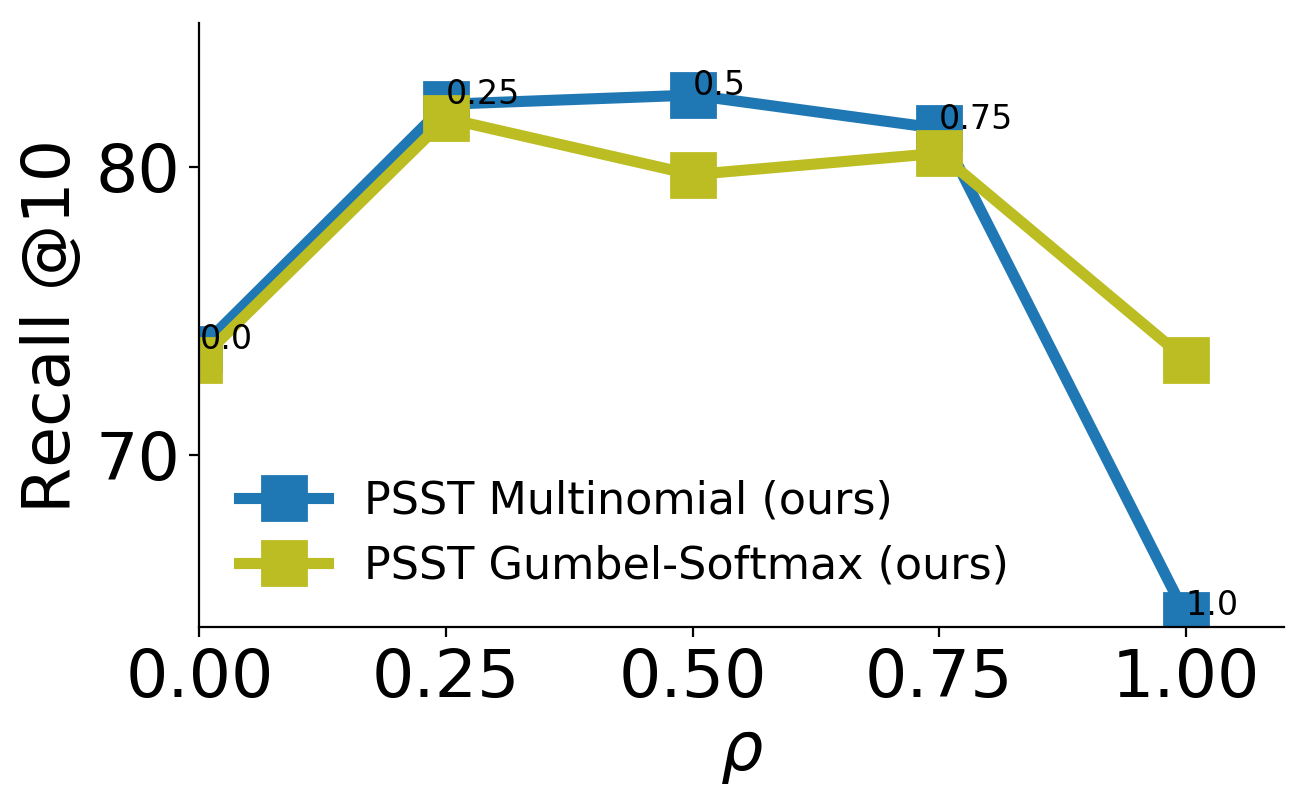}
    \caption{\textit{\textbf{The effect of partial sampling rate ($\rho$) on recall rate}.  Recall was extracted from \figref{fig:recall_vs_cider}, by interpolation, and selecting \cider{} scores at 1.22. Partial sampling ($0<\rho<1$) is better than no sampling ($\rho=1$) or full sampling ($\rho=0$), and results are robust for $\rho \in [0.25, 0.75]$.} }
    \label{fig:recall_vs_rho}
\end{figure}

\figref{fig:recall_vs_rho} illustrates the effect of $\rho$, which controls the sampling ratio in \PSST{}, for \textit{\PSMS{}} and \textit{\PSGSM}.  For a fair comparisons, we fixed the CIDEr score at a given value (the maximal value that overlaps all variants), and report the recall@10 on the interpolated discriminability-naturalness curve of \figref{fig:recall_vs_cider}.  For both methods models with $\rho$ = 0 or 1 gave significantly lower results then models with $\rho$ between 0.25 to 0.75. This is consistent with the idea that using $\rho<1$ (some sampling) is necessary for exposing the listener to sparse inputs, so it does not suffer a catastrophic domain shift at test time.  

\subsection{Ablation Experiments}
\label{sec:Ablation}
To understand the contribution of the different components of our approach, we carried ablation experiments that quantify the benefits of joint-training and partial sampling. Specifically, we evaluated the effect of training only the listener while keeping the speaker model fixed after being pre-trained either using MLE or using the Luo \etal{} baseline. 
In both cases below, since the speaker is not trained, in practice we only optimize $l^{disc}$, without optimizing $l^{nat}$ 

\textbf{Ablation 1: Frozen speaker, Luo}.  Starting from the model of \cite{luo2018discriminability}, we froze the speaker and trained the listener for another 150 epochs. 

\textbf{Ablation 2: Frozen speaker, MLE}. After standard pre-training (speaker 200 epochs, listener 30 epochs), the speaker was kept frozen and the listener was trained on its generated captions (150 epochs, as with all other speakers). 

\textbf{Joint training baseline: Reinforce}. Joint training using the \reinforce{} algorithm (as in Table 1), the method used for training  \cite{luo2018discriminability}.

Table \ref{table:recall_ablation} compares these methods. The top three rows correspond to three models with a frozen network, and the fourth row correspond to joint training using \reinforce. Several results worth discussing.
First, as before, joint training is better than keeping a frozen speaker. Second, 
\cite{luo2018discriminability} trained the speaker with a frozen listener, and  \textit{Frozen Speaker Luo}, improves over that approach by tuning the listener parameters to the Frozen Luo speaker. Third, the \textit{Frozen speaker (MLE)}, which was never adapted based on  listener loss, yields the lowest recall results, as expected. 

\begin{table*}[t]
    \begin{center}
        \begin{tabular}{llrrr}
        {CIDEr=1.13}& ~  &   R@1 &   R@5 &  R@10     \\
        \midrule
        Luo et al. 2018         & ~    & 27.7 &  60.1 &  74.6 \\
        Frozen Speaker (Luo)    & ~    & 32.6 &  67.8 &  81.8 \\
        Frozen Speaker (MLE)    & ~    & 19.3 &  46.9 &  61.3 \\
        \reinforce{}            & ~    & 38.9 &  74.0 &  86.2 \\
        \bottomrule
        \end{tabular}
    \end{center}
    \vspace{5pt}
    \caption{\textit{\textbf{Ablation study}, comparing recall of ``separate training" baselines}. 
    \textit{Recall metrics are reported at  comparable operating points on the discriminative-vs-natural curves. Namely, at points that have the same value of \cider{}=1.13. }
    }
    \label{table:recall_ablation}
\end{table*}


\subsection{Qualitative results}
To get better insight into the captions created by our system, we compare their quality in several ways. 

\begin{figure}[h]
    \begin{center}
    \centering
    \hspace*{-.05cm}
        \includegraphics[width=0.6\linewidth]{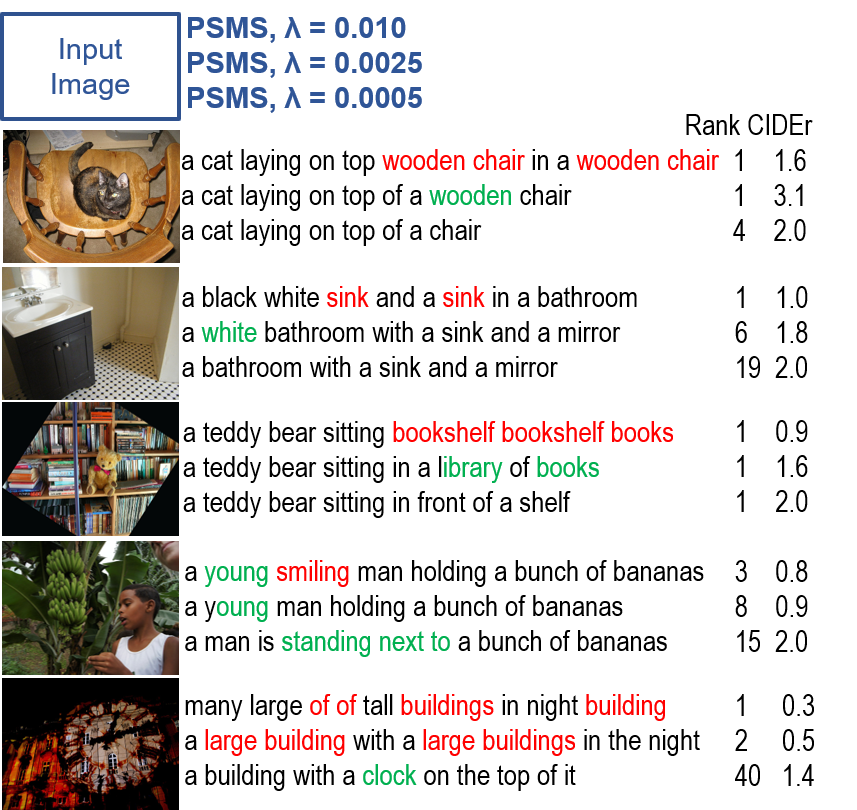}
    \end{center}
   \caption{\textit{\textbf{The effect of the trade-off parameter $\lambda$ on discriminability and naturalness}. All captions were created using \PSMS{}, but with varying values of $\lambda$. The top caption (high $\lambda$), yields more discriminative captions; lower captions (low $\lambda$), leads to more natural sentences. Red text highlight problematic wording; green text highlights correct grammar with additional discriminative information.}}
    \label{fig:qualitative}
\end{figure}

First, we study the effect of the trade-off parameter $\lambda$ on generated captions, balancing discriminability and naturalness. \figref{fig:qualitative} illustrates this. All captions were generated using \PSMS{} ($\rho=0.25$) but trained with different values of $\lambda$. $\lambda$ controls the trade-off between discriminative power and naturalness of captions.

\begin{figure}[h!]
    \centering
    \includegraphics[width=0.98\linewidth]{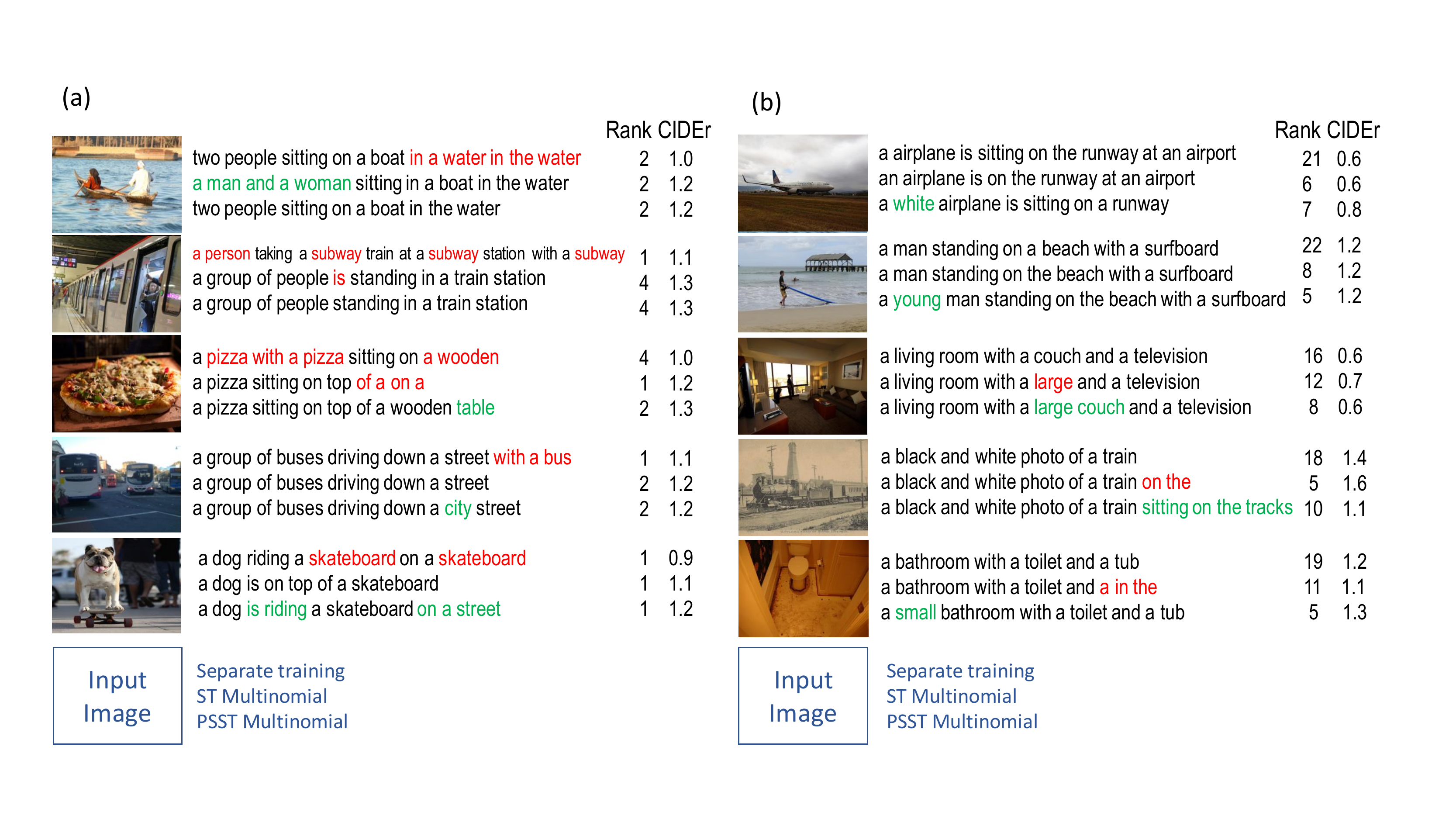}
    \vspace{-30pt}
   \caption{\textit{
   \textbf{The effect of optimization method on naturalness and discriminability}.  Captions were created using three optimization methods, Separate training (listener trained with \reinforce{}) (top), ST multinomial (middle) and \PSMS{} (bottom). (a) naturalness for similar recall values. (b) Discriminability for similar CIDEr values. See text for analysis.}}
   \label{fig:qualitative_fix_recall}
\end{figure}

We then turned to compare \PSMS{} to \STmulti, which can be viewed as the extreme variant of \PSMS{} (always sampling). 
We compare methods in two ways: once by fixing recall and comparing CIDEr scores, and vice versa. \figref{fig:qualitative_fix_recall} illustrate the superior performance of models with better recall-cider curves. The left panel, \figref{fig:qualitative_fix_recall}a compare naturalness for similar recall values of different methods. For each method, $\lambda$ chosen to produce a similar recall@10 rate $\approx 80\%$ (a vertical line in Figure \ref{fig:recall_vs_cider}), yielding mean \cider{} scores of $1.017$, $1.209$, $1.229$ respectively. These examples demonstrate that for this fixed recall, models with higher \cider{} tend to produce more natural captions. To provide "typical" images, we selected images with captions whose CIDEr scores was close to the mean CIDEr of each method, and at the same time where ranked high. Low CIDEr scores are often due to repetitions, and missing nouns. 

The right panel \figref{fig:qualitative_fix_recall}b compares discriminability for similar values of CIDEr.  For each method, $\lambda$ chosen to produce CIDEr $\approx 1.2$, yielding an average image retrieval rank of 20, 9, 8 respectively. The examples demonstrate that for a fixed CIDEr score, models with better image retrieval rank tend to produce more discriminative captions.


\subsection{Human evaluations}
We evaluated the discriminability and the naturalness of various models in a 2-alternative-forced-choice experiment with Amazon Mechanical Turk raters. 

Table \ref{tab:disc} compares models that shared a similar automated naturalness, estimated by selecting models from \figref{fig:recall_vs_cider} with a cider score of $\tildeapprox 1.23$. 
%
%
Raters from the Amazon Mechanical Turk system (AMT), were presented with generated caption of the tested model along with a couple of images: the \textit{correct image}, from which the caption was generated, and a second \textit{distractor image}, which is similar to the "correct" one (and selected by  \cite{luo2018discriminability}). Raters were then asked to choose which of the images is best described by the given caption. This task was designed to measure caption discriminative power, regardless of its naturalness, hence we compared models having similar CIDEr but varying recall@10 levels.  For each method, we tested  the model trained with the smallest $\lambda$ (most natural descriptions). It suggests that \PSMS{} allows raters to achieve slightly better accuracy for discriminating an image from a simlar image. 

Table \ref{tab:nat} further compares three models in terms of their naturalness. The compared models were selected as having similar discriminability. Raters found that \PSMS{} captions were significantly more natural than the reference model and competing models. 
\begin{table}[t]
    \begin{center}
    \begin{tabular}{lc}
        Method & Accuracy \\
        \midrule
        Luo 2018 & 72\% \\
        ST Multinomial & 69\% \\
        \PSMS & \textbf{75\%} \\
        \midrule
        Human & 85\% \\
        \bottomrule
    \end{tabular}
    \end{center}
  \caption{\textit{{\bf Human rater evaluation}: Discriminative power of captions. Reported are accuracy of the majority votes among 5 raters over 300 images.}}
  \label{tab:disc}
\end{table}

\begin{table}[!ht]
    \begin{center}
      \begin{small}\begin{sc}
      \scalebox{0.85}{ 
        \renewcommand{\arraystretch}{0.8} %
        \begin{tabular}{lcc}
        \toprule
        Method & Naturalness &  \\
        \midrule
        {\textsc{PSMS}} & \textbf{+20\%}& \\
        {\textsc{ST Multinomial}} & 0.00 & (reference) \\
        {\textsc{Luo 2018}} & -9\%& \\
        \bottomrule
        \end{tabular}
        }
      \end{sc}\end{small}
    \end{center}
  \caption{{\bf Naturalness of captions}:  
  \textit{Raters were provided with two captions, one from each model, and a single image that the caption describes. They were asked to select the caption that is has proper natural English while also describing the image. Specifically, they were instructed to pay attention to incoherent singular-plural terms, repeating terms and broken sentences. 
  We used ST multinomial as a reference line. All three models were selected to have comparable discriminability, specifically, a recall@10 of $\approx 80\%$ selected from \figref{fig:recall_vs_cider}.}}
  \label{tab:nat}
\end{table}

\begin{table}[h]
    \begin{center}
    \begin{tabular}{lrrrrr}
        Recall@5=90\% & CIDEr & BLEU4 & METOR & ROUGE & SPICE \\
        \midrule
        Luo et al    &      - &      - &      - &      - & - \\ 
        \reinforce{} &  0.431 &  0.173 &  0.188 &  0.435 &  \bf{0.129} \\
        ST-Gumbel SM &  0.484 &  \bf{0.213} &  0.188 &  \bf{0.455} &  0.125 \\
        ST Multinomial &  0.478 &  0.212 &  0.188 &  0.452 &  0.124 \\
        \midrule
        PSST Gumbel Softmax (ours) &  0.485 &  0.207 &  0.188 &  0.447 &  0.126 \\
        PSST Multinomial (ours) &  \bf{0.488} &  \bf{0.213} &  \bf{0.190} &  0.448 &  \bf{0.129} \\
        \bottomrule
    \end{tabular}
    \end{center}
    \vspace{5pt}
    \caption{\textit{\textbf{Evaluation on Flickr30K}. Showed are language quality metrics {\sc{cider, blue4, meteor, rouge}} and {\sc{spice}} for a fixed value of R@5. As in Table \ref{table:coco_metrics}}}
    \label{table:flickr}
\end{table}

\subsection{Evaluations on Flickr30}
We also repeated evaluations on the Flickr30K dataset. This dataset was never used during the development of the method, and evaluations were only computed after all the experiments on COCO were completed. 
\tableref{table:flickr} lists the naturalness metrics for a fixed recall (90\%) on this dataset. The method of \cite{luo2018discriminability} did not reach 90\% recall. Compared to the other approaches for joint training, \PSST{} achieves small improvement in most of the language metrics.

\section{Conclusion}
This paper addresses the problem of building deep models  that can communicate with each-other about perceived world using plain language that is interpretable by people. We find that training jointly can improve both the discriminatibility of captions and their naturalness, compared to separate training. Furthermore, we describe a mechanisms to improve over existing techniques for joint training.  

We describe an approach aimed to handle two challenges of training a speaker and listener jointly. First, we keep the language natural, by restricting it to be similar to a set of captions that were collected in advance in a non-discriminative way. We find that this is sufficient for keeping the language natural, but allowing captions to become much more discriminative. 

Second, we introduce a variant to optimization through a stochastic layer. Since in our architecture, the speaker and listener  cooperate to achieve a shared goal, we find that replacing the sampling procedure for a fraction of the time, and deterministically passing the full distribution provides additional improvement in caption quality. This partial sampling strikes a balance between two effects. First, it passes more information for every sample, reducing the variant of gradient estimate and reaching  better minima of the loss. Second, by providing some low level of sampling, it ensures that networks experience some captions during training time that have the same characteristics like the captions observed at test time, and by that avoid a catastrophic domain shift. 

This work can be extended in several natural ways. First, by allowing the speaker network and listener network to communicate across several rounds (visual dialogues). Second by extending the discriminability measures to more useful metrics of understanding. Finally, by introducing communication between more agents, in scenarios where multi-agent cooperation can be beneficial.

\section*{Acknowledgement}
We deeply thank Ruotian Luo, Gregory Shakhnarovich, Yoav Goldberg and the remaining authors of \citep{luo2018discriminability} for providing and maintaining an implementation of their approach, and for numerous insightful discussions. This research was supported by an Israel science foundation grant 737/18.

\newpage
\bibliographystyle{apalike2}

\newpage
\appendix
\appendixtitleon
\appendixtitletocon
\begin{appendices}

\section{Network architectures and Image features}
Following the discussion in Section \ref{subsec:implementation-details} of the main paper, we expand  the description of model architecture and image features.

\noindent\textbf{The listener network} receives four inputs: the caption,  target image, distractor captions, distractor images.
Every image is represented as a dense vector of size 2048 which is the output of a ResNet 101 \citep{he2016deep}, and is mapped by the network to dimension 1024. Every caption is represented as a sequence of words, each as a vector that has the dimension of the vocabulary. These terms are first mapped to dimension 512, and then enter a GRU with a dimension of 1024. The listener computes the cosine similarity between the output of the GRU and the representation of the image. 

\noindent\textbf{The Speaker network} receives an image, represented using a set of Faster RCNN descriptor. Each descriptor is of size 2048, and is mapped into dimension 512. All descriptors are pooled in a weighted way (attention) and the result is fed to an LSTM. 

\subsection{Pretraining}
The hyper parameters used for pretraining the network, discussed in Section \ref{subsec:implementation-details} in the main paper are from \cite{luo2018code} (learning rate = 5e-4, decay = 0.8 every 15 epochs).

\subsection{Hyper Parameters}
Following Section \ref{subsec:implementation-details} in the main paper, here we explain in details how the hyper parameters were chosen. All hyper parameters were selected using the validation set. 

For \textit{ST Multinomial} and \textit{ST Gumbel-Softmax} we tested learning-rate in (\{1e-4, 5e-4, 1e-3, 5e-3\}) and decay in (\{0.7, 0.75, 0.8\}). For \reinforce{}, we tested learning-rate in (\{5e-4, 5e-3, 1e-2, 5e-2\}) and decay in (\{0.7, 0.8, 0.9\}). For \PSGSM{} and for \PSMS{} we used the same hyperparameters that were found to be best for \textit{ST Multinomial} and \textit{ST Gumbel-Softmax}, and we did not tuned them further (learning rate of $5e-3$ and decay $0.75$).

To select the best learning rate within the above sets, we trained a full curve of recall-vs-cider for each value and selected the rate that maximized the recall@10 for CIDEr=1.125. We repeated the same procedure for the decay (\{0.7, 0.75, 0.8\}), together yielding the best rate = $5e-3$, and the best decay = $0.75$. For \STGSM{} we tested temperatures of $\tau \in \{0.5, 1, 3, 5, 7 ,10\}$, and found $\tau = 1$ was best. 

For \PSGSM{} and \PSMS{} we tested $\rho$ in \{0, 0.25, 0.5, 0.75, 1\}, and show performance as a function of $\rho$ in  \figref{fig:recall_vs_rho}. When training with \reinforce{}, we tested several baselines for reducing variance of the estimator (greedy, ground truth, no baseline) and report results obtained with the ground truth baseline which worked best.
At  test time, we used a beam search of size 2 for generating captions, as in \citep{luo2018discriminability}. 

\section{Human evaluation experiment protocols} 
As discussed  in Section \ref{sec:amt} of the main paper, we used raters from Amazon Mechanical Turk (AMT), to  evaluate the various methods under two experimental setups: (1) We evaluated the \textit{discriminative power of captions}, in an "Image retrieval" experiment and evaluated the \textit{naturalness of captions} in a "caption retrieval" experiment. The results of those experiments are shown in Table \ref{tab:disc} and in Table \ref{tab:nat} respectively. Below are the details of the experimental protocols.

\subsection{Image retrieval experiment}
This experiment contained 300 tasks (HITs). In each task, raters were asked to read a caption generated by the evaluated model, and select one of two images: a "correct" image and a "distractor" one. Pairs were chosen in accordance with \cite{luo2018discriminability}. Each comparison was given to 5 different raters and we report the average accuracy in Table \ref{tab:disc} of the main paper. To verify that raters were skillful, they had to successfully pass a qualification test that contained 10 example tasks. See an example screen shot of the task in \figref{fig:amt_screen_shot} below. 

Raters were given the following instructions: 
\begin{enumerate}
    \setlength{\itemsep}{0.mm}
    \item You are given two images and a sentence caption that should describe one of them. Select the image that is best described by the sentence.
    \item If both images are suitable, choose the one that fits best in your opinion.
    \item Some of the sentences may have incorrect grammar and "broken English" - that's OK! Choose the most suitable image anyway.
\end{enumerate}
  
\subsection{Caption retrieval experiment}
This experiment contained 250-350 tasks (depends on compared model). In each task, raters were asked to look at an image and to choose one of two generated captions: A reference caption from ST MULTINOMIAL model and a caption from an evaluated model. The two evaluated models were PSMS and LOU 2018. In a similar manner to the image retrieval experiment, we gave each comparison to 5 different raters, and averaged the results. The results shown In Table \ref{tab:nat} in the main paper. a qualification test was given to the raters in order to participant in the experiment. See an example screen shot of the task in Figure \figref{fig:amt_screen_shot} below.

Raters were given the following instructions: 
\begin{enumerate}
    \setlength{\itemsep}{0.mm}
    \item Read the two given captions (sentences).
    \item Choose the one that, in your opinion, is better written.
    \item Pay attention to issues like: grammar, singular vs plural, unnecessary repetitions and content coherence.
    \item Sometimes, both sentences (captions) may look like "broken English", that's OK! Choose the one you think is best anyway.
\end{enumerate}

\begin{figure}[t]
    \begin{center}
    \centering
    \hspace*{-.05cm}
    \includegraphics[width=0.45\linewidth, trim={7.8cm 0.1cm 0.1cm, 0.1cm},clip]{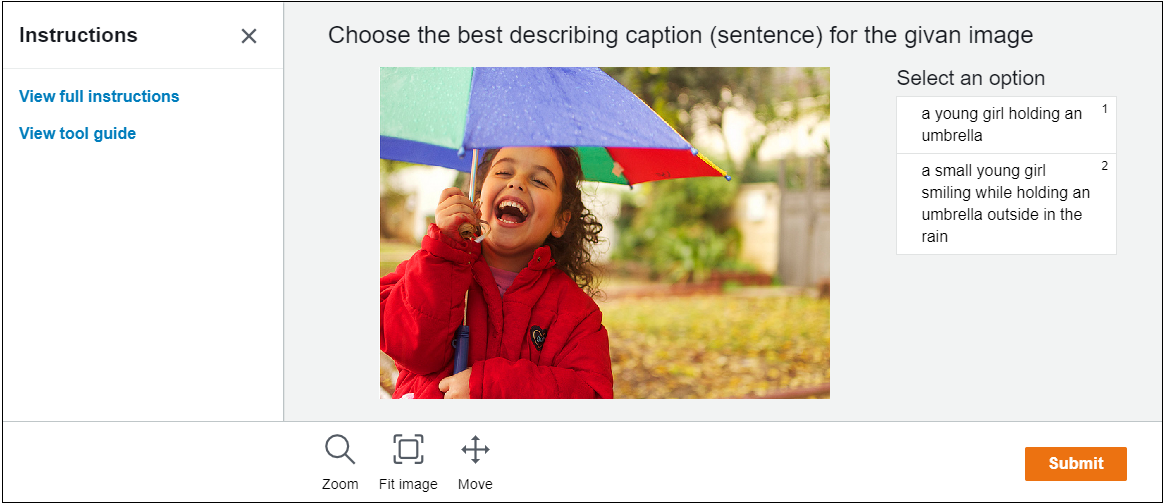}
    \includegraphics[width=0.45\linewidth, trim={0cm 0cm 0cm, 0cm},clip]{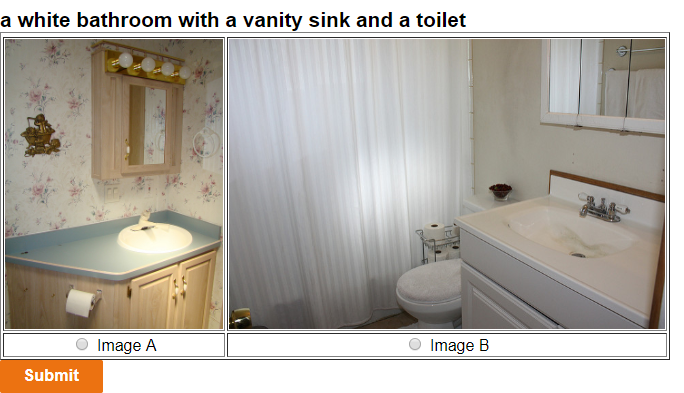}
    \end{center}
    \hspace*{-.5cm}
    \caption{\textbf{Left:} A screen shot of the caption retrieval AMT experiment. \textbf{Right:} A screen shot of the image retrieval AMT experiment.} \label{fig:2_images_1_caption_example}
   \label{fig:amt_screen_shot}
\end{figure}

\end{appendices}

\end{document}